\documentclass[pdflatex,sn-mathphys-num]{sn-jnl}

\usepackage{graphicx}
\usepackage{multirow}
\usepackage{amsmath,amssymb,amsfonts}
\usepackage{amsthm}
\usepackage{mathrsfs}
\usepackage[title]{appendix}
\usepackage{xcolor}
\usepackage{textcomp}
\usepackage{manyfoot}
\usepackage{booktabs}
\usepackage{listings}

\usepackage{algorithm}
\usepackage{algpseudocode}

\theoremstyle{thmstyleone}
\newtheorem{theorem}{Theorem}[section]
\newtheorem{proposition}[theorem]{Proposition}

\theoremstyle{thmstyletwo}

\theoremstyle{thmstylethree}

\newtheorem{assumption}[theorem]{Assumption}

\begin{document}

\title[Article Title]{Uncertainty-Aware Abstention in Large Language Models with Provable Alignment Guarantees}


\author*[1]{Sijin Dong}\email{dong8529278@gmail.com}

\author[1]{Hiroyuki Shinnou}

\affil[1]{\orgname{Ibaraki University}, \country{Japan}}


\abstract{Large language models (LLMs) are increasingly used in question answering (QA) systems, but they can still produce hallucinated or misaligned responses without reliable confidence estimates. Uncertainty quantification (UQ) enables selective answering, where a system responds only when its prediction appears reliable and abstains otherwise. However, uncertainty scores alone are often heuristic, and thresholding them does not provide statistical guarantees on the error rate among accepted answers. 
We propose \textbf{CIC}, a confidence-interval-based calibration framework that converts arbitrary uncertainty scores into risk-controlled selective answering rules. Using a held-out calibration set, CIC assigns each generated response an uncertainty score and a binary error label based on an application-specific alignment criterion. For each candidate threshold, CIC estimates the error rate among accepted answers and constructs a high-probability upper confidence bound using Hoeffding-style or Clopper--Pearson intervals. It then selects the threshold with the highest answering rate whose upper bound remains below a user-specified risk level $\alpha$. 
Under exchangeability, CIC guarantees with probability at least $1-\delta$ that the selected non-null threshold controls the accepted-answer error rate at level $\alpha$. Experiments on closed-ended and open-ended QA benchmarks across seven LLMs and multiple uncertainty estimators show that CIC achieves valid risk control while maintaining strong answering efficiency. These results demonstrate that CIC provides a practical and statistically grounded mechanism for reliable LLM deployment in QA workflows.}




\maketitle

\section{Introduction}

Large language models (LLMs) have achieved remarkable progress in natural language understanding and generation~\citep{wang2026large,jiang2026survey}, and are increasingly deployed as question answering (QA) systems, conversational assistants~\citep{eldem2026development}, and decision-support tools~\citep{song2026decision}. 
Despite their strong empirical performance, LLMs can still generate hallucinated, unsupported, or semantically misaligned responses~\citep{bi2025cotkineticstheoreticalmodelingassessing, huang2025loongsynthesizelongchainofthoughts, wan2025magicwordssharpnessawareprompt, tian2025reinforcementmidtraining, rong2025backdoor,wang2025ascd}. This issue is particularly problematic in reliability-sensitive applications, where an incorrect answer may be more harmful than no answer at all. Therefore, beyond improving the average accuracy of LLMs, it is crucial to develop principled mechanisms that determine when a model output should be trusted and when the system should abstain~\citep{bi-etal-2025-llava, bi2025prismselfpruningintrinsicselection, jiang2025minedprobingupdatingmultimodal, jiang2025koreenhancingknowledgeinjection, zhang2023spot, peng2025visualinputcompressedvisual, li2026graphsubstratedatamodalities, tian2025reinforcementmidtraining}.

Uncertainty quantification (UQ) provides a natural route toward this goal~\citep{gawlikowski2023survey}. Given a query and a generated response, an uncertainty estimator assigns a score intended to reflect the reliability of the output. Such scores can be used for selective answering: the system accepts a response when its uncertainty is sufficiently low and abstains otherwise. A wide range of uncertainty signals have been proposed for LLMs, including entropy-based scores, sampling-based disagreement measures, and semantic-consistency metrics~\citep{duan2024shifting,kuhn2023semantic,wang2025word,wang2024conu}. However, these scores are often heuristic. They may correlate with correctness on average, but they do not perfectly separate correct from incorrect answers. As a result, choosing an uncertainty threshold by intuition or empirical accuracy alone does not provide a statistical guarantee on the error rate among the answers that the system actually returns~\citep{gui2024conformal,Wang_Duan_Wang_Zhu_Chen_Shi_Xu_2026,wang2025lec}.

This limitation motivates the central question of this work: given an arbitrary uncertainty score for an LLM output, can we calibrate a test-time answering rule that accepts as many responses as possible while provably controlling the risk among accepted answers? This question differs from standard accuracy evaluation, which measures model performance over all test samples, regardless of whether the system chooses to answer. It also differs from classical split conformal prediction~\citep{angelopoulos2021uncertainty,angelopoulos2023conformal,angelopoulos2024theoretical}, which typically constructs prediction sets with coverage guarantees. In many QA deployments, however, the system is expected to return a single answer when confident and abstain otherwise~\citep{gui2024conformal}. The relevant reliability target is therefore not set coverage~\citep{wang2024conu,wang2025sconu,wang2025sample,li2026set}, but the error rate conditioned on acceptance.

\begin{figure*}[!t]
    \centering
    \includegraphics[width=\linewidth]{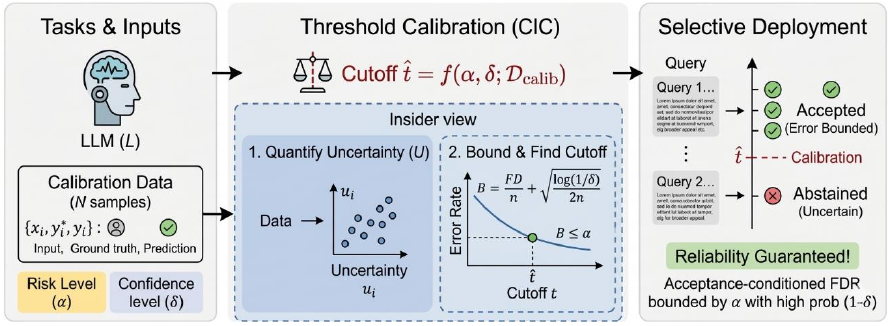}
    \caption{Overview of the CIC framework for risk-controlled selective answering. CIC uses a held-out calibration set to associate each LLM response with an uncertainty score and an alignment label. It then scans candidate uncertainty thresholds, computes an upper confidence bound on the error rate among accepted responses, and selects the largest threshold satisfying the target risk level $\alpha$. During deployment, the calibrated threshold acts as an uncertainty gate: responses below the threshold are accepted, while high-uncertainty responses are abstained from, yielding finite-sample risk control under exchangeability.}
    \label{fig:framework}
\end{figure*}

In this paper, we propose \textbf{CIC}, a confidence-interval-based calibration framework for risk-controlled selective answering with LLMs. We consider a frozen LLM equipped with any user-chosen uncertainty estimator. On a held-out calibration set, each generated response is evaluated by an application-specific alignment criterion, producing a binary label that indicates whether the response is aligned with the ground truth. CIC then links the uncertainty score of each response to its empirical correctness and learns an uncertainty threshold for deployment. 
Figure~\ref{fig:framework} provides an overview of the proposed calibration and selective deployment pipeline.

The key idea is simple: for any candidate threshold, the system accepts calibration examples whose uncertainty scores fall below the threshold, and estimates the error rate among these accepted examples. Since this empirical error rate is computed from finite calibration data, directly using it may lead to unreliable threshold choices. CIC therefore constructs a high-probability upper confidence bound on the threshold-conditioned error rate, using either a Hoeffding-style confidence bound or an exact Clopper--Pearson confidence interval~\citep{clopper1934use}. The final threshold is chosen as the largest candidate threshold whose upper confidence bound does not exceed a user-specified risk level $\alpha$. This choice maximizes the answering rate among certified thresholds while maintaining a finite-sample safety constraint.

At deployment time, the calibrated threshold acts as an uncertainty gate. For a new query, the LLM generates an answer and the uncertainty estimator computes its score. The system returns the answer only if the score is below the calibrated threshold; otherwise, it abstains. Under an exchangeability assumption between calibration and deployment data, CIC guarantees that, with probability at least $1-\delta$, the selected threshold, if non-null, controls the acceptance-conditioned error rate at the target level $\alpha$. In the one-answer-per-query setting, this quantity is equivalent to the marginal false discovery rate of the selective answering policy.

We evaluate CIC on both closed-ended and open-ended QA benchmarks across multiple LLMs and uncertainty estimators. The experimental results show that CIC consistently controls the error rate among accepted answers across a range of user-specified risk levels, while preserving strong answering efficiency. These results demonstrate that confidence-interval calibration can serve as a practical statistical layer on top of heuristic LLM uncertainty scores, enabling more reliable deployment of LLM-based QA systems.

Our main contributions are summarized as follows:
\begin{itemize}
    \item We formulate risk-controlled selective answering for LLM-based QA as the problem of controlling the acceptance-conditioned error rate under an uncertainty-thresholding policy.
    \item We propose CIC, a confidence-interval-based calibration framework that converts arbitrary uncertainty scores into statistically certified abstention rules.
    \item We establish finite-sample, high-probability guarantees for the calibrated threshold under exchangeability, with both Hoeffding-style and Clopper--Pearson upper confidence bounds.
    \item We empirically validate CIC on closed-ended and open-ended QA benchmarks across multiple LLMs, showing reliable risk control with strong answering efficiency.
\end{itemize}

\section{Related Work}
\label{sec: Related Work}

\paragraph{Uncertainty Quantification.}
Uncertainty quantification (UQ) aims to assess the reliability of model predictions and provide users with signals indicating when a model output is likely to be incorrect~\citep{gawlikowski2023survey}. 
For LLM-based generation, recent studies have developed various uncertainty estimators, including semantic entropy (SE)~\citep{kuhn2023semantic}, shift attention to relevance (SAR)~\citep{duan2024shifting}, and word-sequence entropy (WSE)~\citep{wang2025word}. 
These methods provide useful ranking signals for identifying unreliable responses and can be naturally used for selective answering. 
However, LLM uncertainty estimates remain largely heuristic: hallucinated or unsupported responses may still be produced with high confidence, leading to low uncertainty despite being incorrect~\citep{Wang_Duan_Wang_Zhu_Chen_Shi_Xu_2026,wang2025lec}. 
Therefore, directly thresholding such scores does not provide a statistical guarantee on the error rate among accepted answers. 
In contrast, our work treats UQ scores as black-box reliability signals and calibrates them into risk-controlled decision rules with finite-sample guarantees.

\paragraph{Split Conformal Prediction.}
Split conformal prediction (SCP) provides distribution-free and model-agnostic coverage guarantees by calibrating nonconformity scores on a held-out calibration set~\citep{angelopoulos2023conformal,angelopoulos2024theoretical}. 
It has been widely used to construct prediction sets that contain the ground-truth label with user-specified probability, including applications to classification~\citep{angelopoulos2021uncertainty} and segmentation~\citep{tan2025conformal}. 
Recent work has extended conformal methods to LLMs and natural language processing~\citep{campos2024conformal}, including admission coverage~\citep{wang2024conu,wang2025sconu,wang2025sample,quach2024conformal}, conformal factuality~\citep{mohri2024language}, and conformal alignment~\citep{gui2024conformal}. 
These approaches mainly focus on coverage-style guarantees, often through set-valued prediction or conformalized output selection. 
Our setting is different: instead of constructing a prediction set that covers an admissible answer, we study abstention-aware point prediction, where the system either returns a single LLM response or abstains. 
Accordingly, our target is to control the acceptance-conditioned error rate among retained outputs.

\paragraph{Selective Prediction.}
Selective prediction allows a model to abstain from answering when its prediction is considered unreliable, which is especially important in high-stakes or reliability-sensitive applications. 
For LLM-based QA, selective answering can reduce harmful hallucinations by returning responses only when the model is sufficiently confident. 
However, a practical selective system should not merely improve average accuracy after filtering; it should provide a user-specified reliability guarantee for the subset of answers that are actually returned. 
Recent studies have explored abstention and risk control for foundation models, showing that uncertainty-based selection can improve reliability when combined with proper statistical calibration~\citep{wang2025lec,Wang_Duan_Wang_Zhu_Chen_Shi_Xu_2026,jia2026balancerag}. 
Our work follows this direction but focuses on a simple and general confidence-interval-based calibration rule. 
Given any uncertainty score, CIC estimates the threshold-conditioned error rate on a calibration set, constructs an upper confidence bound, and selects the largest threshold satisfying the target risk level. 
This yields a deployable selective answering policy with finite-sample, high-probability control of the error rate among accepted responses.

\section{Methodology}

\subsection{Notation and Problem Formulation}

Let $(X,Y^\star)\sim \mathbb{P}$ denote an input--ground-truth pair, where $X\in\mathcal X$ is a query and $Y^\star\in\mathcal Y$ is its reference answer. We consider a frozen deployed LLM $\mathcal L$, which produces a response
\[
\hat Y = \mathcal L(X;\xi),
\]
where $\xi$ denotes the model's generation randomness; deterministic decoding is a special case.

To evaluate whether the generated answer is acceptable for deployment, we introduce an application-specific alignment criterion
\[
\mathcal A:\mathcal Y\times\mathcal Y\to\{0,1\},
\]
where $\mathcal A(Y^\star,\hat Y)=1$ indicates that the generated answer is aligned with the ground truth, and $\mathcal A(Y^\star,\hat Y)=0$ otherwise. We define the corresponding error indicator as
\begin{equation}
    E = 1-\mathcal A(Y^\star,\hat Y)\in\{0,1\}.
\end{equation}

Given a user-chosen uncertainty measure $\mathcal M$, we compute an uncertainty score
\begin{equation}
    U=\mathcal U(X,\hat Y;\mathcal M)\in\mathbb R,
\end{equation}
where smaller values indicate more reliable predictions. For a threshold $t\in\mathbb R$, we define the thresholding gate
\begin{equation}
    g_t(X,\hat Y)=\mathbf 1\{U\le t\}.
\end{equation}
A prediction is \emph{accepted} when $g_t(X,\hat Y)=1$, and otherwise the system abstains.

Our target is to control the \emph{selection-conditioned error rate} under threshold $t$, defined as
\[
R(t)
:=\mathbb P(E=1\mid U\le t)
=\frac{\mathbb E[E\cdot\mathbf 1\{U\le t\}]}{\mathbb E[\mathbf 1\{U\le t\}]},
\qquad \text{whenever } \mathbb P(U\le t)>0.
\]
This quantity measures the error proportion among accepted predictions. In the one-prediction-per-query selective answering setting, $R(t)$ also coincides with the marginal false discovery rate (mFDR) of the thresholding policy.

Let
\[
\pi(t):=\mathbb P(U\le t)
\]
denote the acceptance probability. The deployment objective is to learn a threshold that accepts as many samples as possible while ensuring
\[
R(t)\le \alpha
\]
for a user-specified risk level $\alpha\in(0,1)$.

We assume access to a held-out calibration set
\[
\mathcal D_{\mathrm{cal}}=\{(X_i,Y_i^\star)\}_{i=1}^N,
\]
which is exchangeable with future deployment data. Running the frozen LLM on $\mathcal D_{\mathrm{cal}}$ yields
\[
\{(X_i,Y_i^\star,\hat Y_i,U_i,E_i)\}_{i=1}^N,
\]
where
\[
\hat Y_i=\mathcal L(X_i;\xi_i),\qquad
U_i=\mathcal U(X_i,\hat Y_i;\mathcal M),\qquad
E_i=1-\mathcal A(Y_i^\star,\hat Y_i).
\]

\subsection{UCB-Based Risk Calibration}

For any threshold $t$, define the calibration acceptance count
\begin{equation}
    n(t)=\sum_{i=1}^N \mathbf 1\{U_i\le t\},
\end{equation}
and the number of false discoveries among accepted samples
\begin{equation}
    FD(t)=\sum_{i=1}^N \mathbf 1\{U_i\le t,\ E_i=1\}.
\end{equation}
The empirical selection-conditioned error rate is
\[
\widehat R(t)=
\begin{cases}
\frac{FD(t)}{n(t)}, & n(t)>0,\\[6pt]
1, & n(t)=0.
\end{cases}
\]

To obtain a finite-sample safety certificate, we construct a one-sided UCB for $R(t)$. We consider two instantiations.

\paragraph{Hoeffding-style UCB.}
For confidence level $1-\gamma$,
\[
B_{\mathrm{H}}(t;\gamma)=
\begin{cases}
\widehat R(t)+\sqrt{\frac{\log(1/\gamma)}{2\,n(t)}}, & n(t)>0,\\[8pt]
1, & n(t)=0.
\end{cases}
\]

\paragraph{Clopper--Pearson-style UCB.}
For confidence level $1-\gamma$,
\[
B_{\mathrm{CP}}(t;\gamma)=
\begin{cases}
\mathrm{BetaInv}\!\left(1-\gamma;\ FD(t)+1,\ n(t)-FD(t)\right), & n(t)>0,\\[6pt]
1, & n(t)=0,
\end{cases}
\]
where $\mathrm{BetaInv}(q;a,b)$ denotes the $q$-quantile of the Beta$(a,b)$ distribution. The Clopper--Pearson bound is exact for binomial proportions and is often less conservative than Hoeffding when the accepted subset is small.

To handle adaptive threshold selection rigorously, we consider a \emph{pre-specified finite threshold grid}
\[
\mathcal T=\{t_1< t_2<\cdots< t_K\},
\]
chosen before observing calibration outcomes. For example, $\mathcal T$ may be a uniform grid over the score range, or a fixed quantile grid obtained from a pilot split.

For each candidate threshold $t\in\mathcal T$, we compute a UCB $B(t;\delta/K)$, where $B$ can be either $B_{\mathrm H}$ or $B_{\mathrm{CP}}$, and $\delta\in(0,1)$ is the target miscoverage probability. We then select the largest threshold whose UCB is below the desired risk level:
\begin{equation}
    \hat t
=
\max\Big\{
t\in\mathcal T:\ B(t;\delta/K)\le \alpha
\Big\}.
\end{equation}
If the feasible set is empty, we return \texttt{NULL}, indicating that the requested risk level is unattainable under the current model and uncertainty signal.

At deployment time, for a new query $X_{\mathrm{test}}$, the model outputs $\hat Y_{\mathrm{test}}$ and uncertainty score $U_{\mathrm{test}}$. The system returns the answer only if
\begin{equation}
    U_{\mathrm{test}}\le \hat t;
\end{equation}
otherwise, it abstains.

\subsection{Framework Summary}

Our framework converts an arbitrary uncertainty score into a statistically certified decision rule for selective answering. The procedure consists of three stages:

First, on a held-out calibration set, we compute for each sample an uncertainty score and an alignment label, thereby linking uncertainty to empirical correctness.

Second, for each candidate threshold, we evaluate the accepted subset induced by that threshold and construct an upper confidence bound on the corresponding selection-conditioned error rate.

Third, we choose the largest threshold whose UCB remains below the user-specified target risk level, and deploy this threshold as a test-time uncertainty gate.

This design preserves the flexibility of heuristic uncertainty estimators while adding a finite-sample safety layer on top of them.

\subsection{Theoretical Guarantees}

We now formalize the validity of the proposed calibration rule.

\begin{assumption}[Exchangeability]
The calibration tuples
\[
\{(U_i,E_i)\}_{i=1}^N
\]
and a future deployment pair $(U_{N+1},E_{N+1})$ are exchangeable. Equivalently, the calibration and deployment data are identically distributed, and the deployed model together with its decoding randomness is fixed across samples.
\end{assumption}

\begin{theorem}[Fixed-threshold validity]
\label{thm:fixed}
Under Assumption 1, fix any threshold $t\in\mathbb R$ such that $\pi(t)=\mathbb P(U\le t)>0$. Let
\[
n(t)=\sum_{i=1}^N \mathbf 1\{U_i\le t\},
\qquad
FD(t)=\sum_{i=1}^N \mathbf 1\{U_i\le t,\ E_i=1\}.
\]
Then conditional on $n(t)=n>0$,
\begin{equation}
    FD(t)\mid n(t)=n \sim \mathrm{Binomial}(n,R(t)).
\end{equation}
Consequently, for any $\gamma\in(0,1)$,
\begin{equation}
    \mathbb P\!\left(R(t)\le B_{\mathrm{CP}}(t;\gamma)\right)\ge 1-\gamma,
\end{equation}
and
\begin{equation}
    \mathbb P\!\left(R(t)\le B_{\mathrm{H}}(t;\gamma)\right)\ge 1-\gamma.
\end{equation}
\end{theorem}

\paragraph{Proof.}
For each calibration point, define three mutually exclusive categories:
\[
C_i=
\begin{cases}
0, & U_i>t \quad\text{(rejected)},\\
1, & U_i\le t,\ E_i=0 \quad\text{(accepted and correct)},\\
2, & U_i\le t,\ E_i=1 \quad\text{(accepted and incorrect)}.
\end{cases}
\]
Because $(U_i,E_i)$ are i.i.d., the category variables $\{C_i\}_{i=1}^N$ are i.i.d. multinomial with probabilities
\[
\mathbb P(C_i=2)=\pi(t)R(t),\qquad
\mathbb P(C_i=1)=\pi(t)(1-R(t)),\qquad
\mathbb P(C_i=0)=1-\pi(t).
\]
Let $N_2=FD(t)$ and $N_1+N_2=n(t)$. Standard multinomial conditioning yields
\[
N_2 \mid (N_1+N_2=n) \sim \mathrm{Binomial}(n,R(t)).
\]
This proves the first claim.

The Clopper--Pearson bound is the exact one-sided $(1-\gamma)$ upper confidence bound for a binomial success probability, hence
\begin{equation}
    \mathbb P\!\left(R(t)\le B_{\mathrm{CP}}(t;\gamma)\mid n(t)=n\right)\ge 1-\gamma
\end{equation}
for every $n>0$, and therefore also unconditionally.

For the Hoeffding bound, conditional on $n(t)=n>0$, the accepted error indicators are i.i.d. Bernoulli with mean $R(t)$. Applying Hoeffding's inequality to their empirical mean $\widehat R(t)$ gives
\begin{equation}
    \mathbb P\!\left(
R(t) >
\widehat R(t)+\sqrt{\frac{\log(1/\gamma)}{2n}}
\;\middle|\; n(t)=n
\right)\le \gamma.
\end{equation}
Again, averaging over $n$ yields the unconditional statement.
\hfill $\square$

\begin{theorem}[Adaptive threshold guarantee]
\label{thm:adaptive}
Under Assumption 1, let $\mathcal T=\{t_1,\dots,t_K\}$ be a fixed finite threshold grid, and let
\begin{equation}
    \hat t
=
\max\Big\{
t\in\mathcal T:\ B(t;\delta/K)\le \alpha
\Big\},
\end{equation}
where $B$ is either $B_{\mathrm H}$ or $B_{\mathrm{CP}}$. Then
\begin{equation}
    \mathbb P\!\left(
\hat t=\texttt{NULL}
\ \text{or}\
R(\hat t)\le \alpha
\right)\ge 1-\delta.
\end{equation}
Equivalently, with probability at least $1-\delta$, every non-null threshold returned by the algorithm satisfies the desired selection-conditioned risk guarantee.
\end{theorem}

\paragraph{Proof.}
For each fixed $t\in\mathcal T$, Theorem~\ref{thm:fixed} gives
\begin{equation}
    \mathbb P\!\left(R(t)\le B(t;\delta/K)\right)\ge 1-\delta/K.
\end{equation}
Define the simultaneous validity event
\begin{equation}
    \mathcal E
=
\bigcap_{t\in\mathcal T}
\left\{
R(t)\le B(t;\delta/K)
\right\}.
\end{equation}
By the union bound,
\begin{equation}
    \mathbb P(\mathcal E)
\ge
1-\sum_{t\in\mathcal T}\frac{\delta}{K}
=
1-\delta.
\end{equation}

On the event $\mathcal E$, every candidate threshold satisfying $B(t;\delta/K)\le \alpha$ also satisfies
\begin{equation}
    R(t)\le B(t;\delta/K)\le \alpha.
\end{equation}
Therefore, if the algorithm returns a non-null threshold $\hat t$, then necessarily
\begin{equation}
    R(\hat t)\le \alpha.
\end{equation}
If no threshold is feasible, the algorithm returns \texttt{NULL}, which is also covered by the theorem statement. Hence
\begin{equation}
    \mathbb P\!\left(
\hat t=\texttt{NULL}
\ \text{or}\
R(\hat t)\le \alpha
\right)\ge 1-\delta.
\end{equation}
\hfill $\square$

\begin{algorithm}[!t]
  \caption{Upper Confidence Bound Computation and Threshold Calibration}
  \label{alg: calibration}
  \begin{algorithmic}[1]
    \State {\bfseries Input:} Deployed LLM $\mathcal{L}$, $N$ calibration data points $\{ ( x_i, y_i^*, \hat{y}_i ) \}_{i=1}^{N}$, uncertainty measure $\mathcal{M}$, alignment criterion $\mathcal{A}$, risk level $\alpha$, significance level $\delta$
    \State \textbf{Output:} An FDR-controlled threshold $\hat{t}$
    \State Quantify the uncertainty score of $\hat{y}_i$ for all $i \in [N]$: $u_i \leftarrow \mathcal{U}(x_i; \mathcal{G})$;
    \State Evaluate whether $\hat{y}_i$ is aliged with $y_i^*$ based on $\mathcal{A}$ for all $i \in [N]$: $a_i \leftarrow \mathbf{1}\{ A(y_i^*, \hat{y}_i) = 1 \}$;
    
    \State Sort uncertainty scores $u_{(1)} \le \cdots \le u_{(N)}$, with corresponding $a_{(1)}, \ldots, a_{(N)}$;
    \State Initialize $\hat{t} \leftarrow $ \texttt{NULL};
    \For{$i = 1$ \textbf{to} $N$}
    \State Let $t \leftarrow u_{(i)}$;
    \State Count the number of accepted answers: $n \leftarrow \sum_{j=1}^{N} \mathbf{1} \{ u_{(j)} \leq t \}$;
    \State Count the number of misaligned answers among all accepted ones (i.e., false discovery): $FD \leftarrow \sum_{j=1}^{N} \mathbf{1} \{ u_{(j)} \leq t \land a_{(j)} = 0 \}$;
    
    \State Compute the Hoeffding-style ($1-\delta$)-upper confidence bound: $\mathrm{B} \leftarrow \frac{FD}{n} 
        + \sqrt{\frac{\log(1/\delta)}{2 n}}$;
    \State \textbf{Remark:} Or compute the Clopper--Pearson-style exact ($1-\delta$)-upper confidence bound: $\mathrm{B} \leftarrow \mathrm{BetaInv}\!\left(1-\delta;\; FD+1,\; n-FD\right)$;
    
    \If{$\mathrm{B} \le \alpha$}
        \State Update $\hat{t} \leftarrow t$;
    \EndIf
\EndFor
\If{$\hat{t} = $ \texttt{NULL}}
    \State \textbf{Return} ``Risk level $\alpha$ is unattainable''
\Else
    \State \textbf{Return} $\hat{t}$ 
\EndIf 

  \end{algorithmic}
\end{algorithm}

\begin{proposition}[Maximal acceptance among certified thresholds]
\label{prop:maxacc}
Let
\begin{equation}
    \mathcal T_{\mathrm{safe}}
=
\{t\in\mathcal T:\ B(t;\delta/K)\le \alpha\}.
\end{equation}
If $\mathcal T_{\mathrm{safe}}$ is non-empty, then the selected threshold
\begin{equation}
    \hat t = \max \mathcal T_{\mathrm{safe}}
\end{equation}
maximizes the acceptance probability $\pi(t)$ over all certified candidate thresholds:
\begin{equation}
    \pi(\hat t)=\max_{t\in\mathcal T_{\mathrm{safe}}}\pi(t).
\end{equation}
\end{proposition}

\paragraph{Proof.}
For any $t_1\le t_2$, we have
\begin{equation}
    \mathbf 1\{U\le t_1\}\le \mathbf 1\{U\le t_2\},
\end{equation}
hence $\pi(t_1)\le \pi(t_2)$. Therefore, among all thresholds in $\mathcal T_{\mathrm{safe}}$, the largest one has the largest acceptance probability.
\hfill $\square$

\subsection{Practical Remarks}

\paragraph{Choice of UCB.}
The Hoeffding-style bound is closed-form and simple, while the Clopper--Pearson bound is exact for binomial proportions and is generally preferable when the accepted subset is small. In practice, both can be implemented with the same calibration pipeline.

\paragraph{Unattainable risk levels.}
If no threshold satisfies the UCB constraint, the target risk level $\alpha$ is unattainable for the current model--uncertainty pair at confidence level $1-\delta$. This is an intrinsic limitation of the underlying model quality and uncertainty discriminability, rather than a failure of the calibration procedure itself.

\paragraph{On implementation with empirical score grids.}
For theoretical strictness, the candidate threshold grid should be fixed in advance. In practice, one may scan all unique calibration scores for finer resolution; however, the exact guarantee in Theorem~\ref{thm:adaptive} is most cleanly stated for a pre-specified finite grid, or alternatively with an additional split / simultaneous testing correction.

\paragraph{Interpretation.}
Our method does not require the uncertainty score itself to be calibrated. Instead, it treats the uncertainty estimator as a ranking signal, and calibrates a statistically valid operating threshold on top of it. In this sense, the proposed framework converts arbitrary heuristic uncertainty into a deployable abstention policy with finite-sample reliability guarantees.

\paragraph{Implementation.} 
Since the selection-conditioned risk is near-monotonic, i.e., it tends to decrease as the threshold increases, we adopt fixed-sequence testing instead of Bonferroni correction (i.e., $\delta$ instead of $\delta/K$)~\citep{jung2025trust,Wang_Duan_Wang_Zhu_Chen_Shi_Xu_2026}. 
The pseudocode for our calibration framework is presented by Algorithm~\ref{alg: calibration}.

\section{Experiments}
\label{sec: Experiments}

\subsection{Experimental Settings}
\label{sec: Experimental Settings}

\paragraph{\emph{Base Models.}}
We evaluate CIC on seven widely used instruction-tuned LLMs with different model families and scales, including Qwen2.5-3B-Instruct, Qwen2.5-7B-Instruct, Qwen2.5-14B-Instruct~\citep{bai2023qwen}, OpenChat-3.5~\citep{wang2024openchat}, Vicuna-7B-v1.5, Vicuna-13B-v1.5~\cite{zheng2023judging}, and LLaMA-3.1-8B-Instruct~\citep{touvron2023llama}. 
These models cover both relatively small and stronger instruction-following LLMs, allowing us to examine whether the proposed calibration framework is robust across different base error rates and uncertainty-quality regimes.

\begin{table*}[!t]
\caption{Test-time FDR (mean $\pm$ std across 100 splits) on CommonsenseQA.}
\label{tab:commonsenseqa_fdr}
\centering
\small
\setlength{\tabcolsep}{4pt}
\renewcommand{\arraystretch}{1.15}
\resizebox{\textwidth}{!}{%
\begin{tabular}{llcc cc cc cc cc c}
\toprule
\multirow{2}{*}{LLMs} & \multirow{2}{*}{UCBs/$\alpha$} &
\multicolumn{2}{c}{0.05} & \multicolumn{2}{c}{0.10} & \multicolumn{2}{c}{0.15} &
\multicolumn{2}{c}{0.20} & \multicolumn{2}{c}{0.25} & \multirow{2}{*}{Base ER} \\
\cmidrule(lr){3-4}\cmidrule(lr){5-6}\cmidrule(lr){7-8}\cmidrule(lr){9-10}\cmidrule(lr){11-12}
& & FDR & Fail & FDR & Fail & FDR & Fail & FDR & Fail & FDR & Fail & \\
\midrule

\multirow{2}{*}{Qwen2.5-3B-Instruct} & UCB-HFD & - & 100 & $0.074 \pm 0.009$ & 0 & $0.128 \pm 0.008$ & 0 & $0.180 \pm 0.009$ & 0 & $0.230 \pm 0.009$ & 0 & \multirow{2}{*}{0.239} \\
& UCB-CP  & $0.040 \pm 0.008$ & 34 & $0.090 \pm 0.008$ & 0 & $0.139 \pm 0.009$ & 0 & $0.188 \pm 0.009$ & 0 & $0.237 \pm 0.007$ & 0 & {}\\
\midrule

\multirow{2}{*}{Qwen2.5-7B-Instruct} & UCB-HFD & $0.024 \pm 0.004$ & 89 & $0.079 \pm 0.008$ & 0 & $0.131 \pm 0.008$ & 0 & $0.183 \pm 0.008$ & 0 & $0.211 \pm 0.004$ & 0 & 0.212 \\
& UCB-CP  & $0.043 \pm 0.006$ & 0  & $0.092 \pm 0.007$ & 0 & $0.142 \pm 0.008$ & 0 & $0.192 \pm 0.008$ & 0 & $0.211 \pm 0.004$ & 0 & 0.212 \\
\midrule

\multirow{2}{*}{OpenChat-3.5} & UCB-HFD & $0.023 \pm 0.006$ & 7 & $0.079 \pm 0.007$ & 0 & $0.131 \pm 0.008$ & 0 & $0.180 \pm 0.005$ & 0 & $0.182 \pm 0.004$ & 0 & 0.182 \\
& UCB-CP  & $0.043 \pm 0.006$ & 0 & $0.092 \pm 0.007$ & 0 & $0.140 \pm 0.008$ & 0 & $0.182 \pm 0.004$ & 0 & $0.182 \pm 0.004$ & 0 & 0.182 \\
\midrule

\multirow{2}{*}{Vicuna-7b-v1.5} & UCB-HFD & - & 100 & - & 100 & $0.121 \pm 0.020$ & 97 & $0.156 \pm 0.021$ & 3 & $0.220 \pm 0.014$ & 0 & 0.410 \\
& UCB-CP  & - & 100 & $0.080 \pm 0.024$ & 66 & $0.113 \pm 0.029$ & 7 & $0.178 \pm 0.016$ & 0 & $0.233 \pm 0.012$ & 0 & 0.410 \\
\midrule

\multirow{2}{*}{LLaMA-3.1-8B-Instruct} & UCB-HFD & - & 100 & $0.074 \pm 0.008$ & 0 & $0.128 \pm 0.008$ & 0 & $0.180 \pm 0.009$ & 0 & $0.231 \pm 0.009$ & 0 & 0.271 \\
& UCB-CP  & $0.041 \pm 0.008$ & 0 & $0.090 \pm 0.008$ & 0 & $0.140 \pm 0.008$ & 0 & $0.190 \pm 0.009$ & 0 & $0.238 \pm 0.008$ & 0 & 0.271 \\
\midrule

\multirow{2}{*}{Vicuna-13b-v1.5} & UCB-HFD & - & 100 & - & 100 & $0.117 \pm 0.014$ & 0 & $0.174 \pm 0.011$ & 0 & $0.227 \pm 0.010$ & 0 & 0.347 \\
& UCB-CP  & $0.055 \pm 0.010$ & 86 & $0.071 \pm 0.023$ & 0 & $0.135 \pm 0.014$ & 0 & $0.185 \pm 0.011$ & 0 & $0.237 \pm 0.010$ & 0 & 0.347 \\
\midrule

\multirow{2}{*}{Qwen2.5-14B-Instruct} & UCB-HFD & $0.023 \pm 0.005$ & 19 & $0.080 \pm 0.006$ & 0 & $0.131 \pm 0.007$ & 0 & $0.168 \pm 0.004$ & 0 & $0.168 \pm 0.004$ & 0 & 0.168 \\
& UCB-CP  & $0.044 \pm 0.006$ & 0  & $0.092 \pm 0.007$ & 0 & $0.141 \pm 0.007$ & 0 & $0.168 \pm 0.004$ & 0 & $0.168 \pm 0.004$ & 0 & 0.168 \\
\bottomrule
\end{tabular}%
}
\end{table*}

\paragraph{\emph{Benchmarks.}}
Following prior work~\citep{wang2025lec,Wang_Duan_Wang_Zhu_Chen_Shi_Xu_2026}, we conduct experiments on two representative QA benchmarks: CommonsenseQA~\citep{talmor-etal-2019-commonsenseqa} and TriviaQA~\citep{joshi-etal-2017-triviaqa}. 
CommonsenseQA is a closed-ended commonsense reasoning benchmark, where each question is associated with a set of candidate answers and the model is expected to select the correct one. 
TriviaQA is an open-ended question answering benchmark that requires models to generate free-form answers based on factual knowledge. 
Together, these two datasets allow us to evaluate CIC under both constrained answer selection and open-ended generation settings. 

\paragraph{\emph{Uncertainty Estimators.}}
We use semantic entropy as the main uncertainty estimator for LLM outputs~\citep{kuhn2023semantic}. 
For each input question, we sample multiple responses from the base LLM and estimate semantic-level uncertainty by measuring the dispersion of generated answers after accounting for semantic equivalence. 
Unless otherwise specified, we use 10 sampled responses per question. 
The resulting uncertainty score is used as the ranking signal for selective answering, where lower uncertainty indicates a more reliable model response.

\paragraph{\emph{Alignment Evaluation.}}
For CommonsenseQA, correctness is evaluated by comparing the model-selected answer with the ground-truth option. 
For TriviaQA, because the model output is free-form, we evaluate whether the generated answer is semantically aligned with the reference answer. 
Following prior work~\citep{duan2024shifting,wang2025word}, we use sentence similarity with a threshold of 0.6 to determine whether a response should be treated as correct. 
The resulting binary correctness label is then converted into an error indicator and used for confidence-interval calibration.

\paragraph{\emph{Calibration Protocol.}}
For each dataset and model, we randomly split the data into calibration and test subsets. 
The calibration set is used only to select the uncertainty threshold, while the test set is used only for evaluation. 
Following prior work, we repeat the calibration-test split 100 times and report the mean and standard deviation over 100 trials. 
By default, we use a calibration-test split ratio of 0.5. 
We evaluate multiple user-specified risk levels, $\alpha \in \{0.05, 0.10, 0.15, 0.20, 0.25\}$.

\begin{table*}[!t]
\centering
\caption{Test-time FDR (mean ± std across 100 splits) at various user-specified risk levels on TriviaQA with seven LLMs, utilizing semantic entropy (10 samples per question). We employ sentence similarity with a threshold of 0.6 for correctness evaluation.}
\label{tab:triviaqa_fdr}
\resizebox{\textwidth}{!}{
\begin{tabular}{llccccc ccccc c}
\toprule
\multirow{2}{*}{LLMs} & \multirow{2}{*}{UCBs/$\alpha$} 
& \multicolumn{2}{c}{0.05} & \multicolumn{2}{c}{0.1} & \multicolumn{2}{c}{0.15} & \multicolumn{2}{c}{0.2} & \multicolumn{2}{c}{0.25} & \multirow{2}{*}{Base ER} \\
\cmidrule(lr){3-4}\cmidrule(lr){5-6}\cmidrule(lr){7-8}\cmidrule(lr){9-10}\cmidrule(lr){11-12}
& & FDR & Fail & FDR & Fail & FDR & Fail & FDR & Fail & FDR & Fail & \\
\midrule

\multirow{2}{*}{Qwen2.5-3B-Instruct}
& UCB-HFD & -- & 100 & 0.064$\pm$0.012 & 0 & 0.124$\pm$0.008 & 0 & 0.163$\pm$0.004 & 0 & 0.163$\pm$0.004 & 0 & 0.163 \\
& UCB-CP  & 0.038$\pm$0.004 & 34 & 0.086$\pm$0.010 & 0 & 0.136$\pm$0.008 & 0 & 0.163$\pm$0.004 & 0 & 0.163$\pm$0.004 & 0 & \\
\midrule

\multirow{2}{*}{Qwen2.5-7B-Instruct}
& UCB-HFD & -- & 100 & 0.077$\pm$0.007 & 0 & 0.125$\pm$0.004 & 0 & 0.125$\pm$0.004 & 0 & 0.125$\pm$0.004 & 0 & 0.125 \\
& UCB-CP  & 0.037$\pm$0.007 & 4 & 0.091$\pm$0.007 & 0 & 0.125$\pm$0.004 & 0 & 0.125$\pm$0.004 & 0 & 0.125$\pm$0.004 & 0 & \\
\midrule

\multirow{2}{*}{OpenChat-3.5}
& UCB-HFD & -- & 100 & 0.078$\pm$0.008 & 0 & 0.118$\pm$0.004 & 0 & 0.118$\pm$0.004 & 0 & 0.118$\pm$0.004 & 0 & 0.118 \\
& UCB-CP  & 0.039$\pm$0.009 & 0 & 0.092$\pm$0.008 & 0 & 0.118$\pm$0.004 & 0 & 0.118$\pm$0.004 & 0 & 0.118$\pm$0.004 & 0 & \\
\midrule

\multirow{2}{*}{Vicuna-7b-v1.5}
& UCB-HFD & -- & 100 & 0.065$\pm$0.012 & 0 & 0.126$\pm$0.009 & 0 & 0.169$\pm$0.004 & 0 & 0.169$\pm$0.004 & 0 & 0.169 \\
& UCB-CP  & 0.057$\pm$0.000 & 99 & 0.086$\pm$0.009 & 0 & 0.138$\pm$0.009 & 0 & 0.169$\pm$0.004 & 0 & 0.169$\pm$0.004 & 0 & \\
\midrule

\multirow{2}{*}{LLaMA-3.1-8B-Instruct}
& UCB-HFD & 0.025$\pm$0.002 & 52 & 0.078$\pm$0.006 & 0 & 0.115$\pm$0.003 & 0 & 0.115$\pm$0.003 & 0 & 0.115$\pm$0.003 & 0 & 0.115 \\
& UCB-CP  & 0.042$\pm$0.006 & 0 & 0.090$\pm$0.007 & 0 & 0.115$\pm$0.003 & 0 & 0.115$\pm$0.003 & 0 & 0.115$\pm$0.003 & 0 & \\
\midrule

\multirow{2}{*}{Vicuna-13b-v1.5}
& UCB-HFD & -- & 100 & 0.077$\pm$0.008 & 0 & 0.116$\pm$0.004 & 0 & 0.116$\pm$0.004 & 0 & 0.116$\pm$0.004 & 0 & 0.116 \\
& UCB-CP  & 0.037$\pm$0.009 & 0 & 0.090$\pm$0.007 & 0 & 0.116$\pm$0.004 & 0 & 0.116$\pm$0.004 & 0 & 0.116$\pm$0.004 & 0 & \\
\midrule

\multirow{2}{*}{Qwen2.5-14B-Instruct}
& UCB-HFD & -- & 100 & 0.079$\pm$0.006 & 0 & 0.088$\pm$0.003 & 0 & 0.088$\pm$0.003 & 0 & 0.088$\pm$0.003 & 0 & 0.088 \\
& UCB-CP  & 0.040$\pm$0.007 & 0 & 0.088$\pm$0.004 & 0 & 0.088$\pm$0.003 & 0 & 0.088$\pm$0.003 & 0 & 0.088$\pm$0.003 & 0 & \\
\bottomrule
\end{tabular}
}
\end{table*}

\paragraph{\emph{Compared UCBs.}}
We instantiate CIC with two upper confidence bounds: a Hoeffding-style UCB, denoted as UCB-HFD, and a Clopper--Pearson-style UCB, denoted as UCB-CP. 
UCB-HFD provides a simple closed-form bound, while UCB-CP gives an exact binomial confidence upper bound and is often more suitable when the accepted subset is small. 
For each risk level, CIC selects the largest threshold whose UCB does not exceed $\alpha$. 
If no threshold satisfies the constraint, CIC returns no feasible threshold; we report the number of such failed calibration trials as ``Fail''.

\paragraph{\emph{Evaluation Metrics.}}
We report two main metrics. 
First, we evaluate the empirical false discovery rate (FDR), defined as the proportion of incorrect responses among accepted outputs on the test set. 
This corresponds to the acceptance-conditioned error rate studied in our theoretical analysis. 
Second, we report power, defined as the proportion of test samples accepted by the calibrated threshold. 
A reliable selective answering system should keep FDR below the target risk level while achieving high power.

\subsection{Empirical Results}

\paragraph{\emph{Risk Control on CommonsenseQA.}}
Table~\ref{tab:commonsenseqa_fdr} reports the test-time FDR on CommonsenseQA under different risk levels. 
Across most models and risk levels, both UCB-HFD and UCB-CP successfully keep the empirical FDR below the user-specified target $\alpha$. 
This confirms that CIC can convert heuristic uncertainty scores into statistically reliable selective answering rules. 
As $\alpha$ increases, the calibrated threshold becomes less conservative, and the observed FDR correspondingly increases toward the target level. 
This trend is expected: a larger risk budget allows the system to accept more uncertain responses, leading to higher answering coverage but also a higher error rate among accepted outputs.

The results also show that the feasibility of strict risk control depends on the underlying model and uncertainty estimator. 
For very small risk levels such as $\alpha=0.05$, some model--UCB combinations fail to find a feasible threshold in a subset of calibration splits. 
This does not indicate a failure of the calibration procedure; rather, it means that the given model and uncertainty signal cannot certify such a strict risk level with the available calibration data. 
In contrast, when the target risk level is moderately relaxed, CIC usually finds feasible thresholds and maintains valid empirical FDR control.

\paragraph{\emph{Risk Control on TriviaQA.}}
Table~\ref{tab:triviaqa_fdr} presents the test-time FDR results on TriviaQA. 
Compared with CommonsenseQA, TriviaQA involves open-ended generation and semantic answer matching, making the evaluation more challenging. 
Nevertheless, CIC still achieves reliable risk control across the evaluated LLMs. 
For most models, the empirical FDR remains below or close to the target risk level across all tested values of $\alpha$. 
This demonstrates that CIC is not limited to closed-ended answer selection, but can also be applied to open-ended QA when an appropriate alignment criterion is available.

\begin{table*}[!t]
\centering
\caption{Test-time Power (mean across 100 splits) at various user-specified risk levels on TriviaQA with seven LLMs, utilizing semantic entropy (10 samples per question).}
\label{tab:triviaqa_power}
\resizebox{\textwidth}{!}{
\begin{tabular}{llccccc ccccc}
\toprule
\multirow{2}{*}{LLMs} & \multirow{2}{*}{UCBs/$\alpha$} 
& \multicolumn{2}{c}{0.05} & \multicolumn{2}{c}{0.1} & \multicolumn{2}{c}{0.15} & \multicolumn{2}{c}{0.2} & \multicolumn{2}{c}{0.25} \\
\cmidrule(lr){3-4}\cmidrule(lr){5-6}\cmidrule(lr){7-8}\cmidrule(lr){9-10}\cmidrule(lr){11-12}
& & Power & Fail & Power & Fail & Power & Fail & Power & Fail & Power & Fail \\
\midrule

\multirow{2}{*}{Qwen2.5-3B-Instruct}
& UCB-HFD & -- & 100 & 0.381 & 0 & 0.769 & 0 & 1.000 & 0 & 1.000 & 0 \\
& UCB-CP  & 0.238 & 34  & 0.517 & 0 & 0.845 & 0 & 1.000 & 0 & 1.000 & 0 \\
\midrule

\multirow{2}{*}{Qwen2.5-7B-Instruct}
& UCB-HFD & -- & 100 & 0.741 & 0 & 0.999 & 0 & 1.000 & 0 & 1.000 & 0 \\
& UCB-CP  & 0.390 & 4   & 0.833 & 0 & 1.000 & 0 & 1.000 & 0 & 1.000 & 0 \\
\midrule

\multirow{2}{*}{OpenChat-3.5}
& UCB-HFD & -- & 100 & 0.889 & 0 & 1.000 & 0 & 1.000 & 0 & 1.000 & 0 \\
& UCB-CP  & 0.670 & 0   & 0.929 & 0 & 1.000 & 0 & 1.000 & 0 & 1.000 & 0 \\
\midrule

\multirow{2}{*}{Vicuna-7b-v1.5}
& UCB-HFD & -- & 100 & 0.425 & 0 & 0.802 & 0 & 1.000 & 0 & 1.000 & 0 \\
& UCB-CP  & 0.299 & 99  & 0.567 & 0 & 0.857 & 0 & 1.000 & 0 & 1.000 & 0 \\
\midrule

\multirow{2}{*}{LLaMA-3.1-8B-Instruct}
& UCB-HFD & 0.534 & 52  & 0.904 & 0 & 1.000 & 0 & 1.000 & 0 & 1.000 & 0 \\
& UCB-CP  & 0.716 & 0   & 0.937 & 0 & 1.000 & 0 & 1.000 & 0 & 1.000 & 0 \\
\midrule

\multirow{2}{*}{Vicuna-13b-v1.5}
& UCB-HFD & -- & 100 & 0.835 & 0 & 1.000 & 0 & 1.000 & 0 & 1.000 & 0 \\
& UCB-CP  & 0.567 & 0   & 0.891 & 0 & 1.000 & 0 & 1.000 & 0 & 1.000 & 0 \\
\midrule

\multirow{2}{*}{Qwen2.5-14B-Instruct}
& UCB-HFD & -- & 100 & 0.970 & 0 & 1.000 & 0 & 1.000 & 0 & 1.000 & 0 \\
& UCB-CP  & 0.621 & 0   & 0.999 & 0 & 1.000 & 0 & 1.000 & 0 & 1.000 & 0 \\
\bottomrule
\end{tabular}
}
\end{table*}

\begin{table*}[!t]
\caption{Test-time Power (mean across 100 splits) at various user-specified risk levels on CommonsenseQA with seven LLMs,}
\label{tab:commonsenseqa_power}
\centering
\small
\setlength{\tabcolsep}{6pt}
\renewcommand{\arraystretch}{1.15}
\resizebox{\textwidth}{!}{%
\begin{tabular}{llccccc}
\toprule
\multirow{2}{*}{LLMs} & \multirow{2}{*}{UCBs/$\alpha$} &
\multicolumn{5}{c}{Risk level ($\alpha$)} \\
\cmidrule(lr){3-7}
& & 0.05 & 0.10 & 0.15 & 0.20 & 0.25 \\
\midrule

\multirow{2}{*}{Qwen2.5-3B-Instruct}
& UCB-HFD & -   & 0.502 & 0.774 & 0.895 & 0.984 \\
& UCB-CP  & 0.176 & 0.623 & 0.805 & 0.909 & 0.995 \\
\midrule

\multirow{2}{*}{Qwen2.5-7B-Instruct}
& UCB-HFD & 0.345 & 0.720 & 0.858 & 0.952 & 1.000 \\
& UCB-CP  & 0.572 & 0.766 & 0.882 & 0.970 & 1.000 \\
\midrule

\multirow{2}{*}{OpenChat-3.5}
& UCB-HFD & 0.488 & 0.802 & 0.921 & 0.999 & 1.000 \\
& UCB-CP  & 0.655 & 0.839 & 0.940 & 1.000 & 1.000 \\
\midrule


\multirow{2}{*}{LLaMA-3.1-8B-Instruct}
& UCB-HFD & -   & 0.551 & 0.740 & 0.860 & 0.941 \\
& UCB-CP  & 0.391 & 0.608 & 0.770 & 0.874 & 0.955 \\
\midrule

\multirow{2}{*}{Vicuna-13b-v1.5}
& UCB-HFD & -   & -   & 0.384 & 0.617 & 0.763 \\
& UCB-CP  & 0.041 & 0.132 & 0.468 & 0.646 & 0.794 \\
\midrule

\multirow{2}{*}{Qwen2.5-14B-Instruct}
& UCB-HFD & 0.446 & 0.833 & 0.942 & 1.000 & 1.000 \\
& UCB-CP  & 0.661 & 0.865 & 0.959 & 1.000 & 1.000 \\
\bottomrule
\end{tabular}%
}
\end{table*}

We further observe that when the base model already has a relatively low error rate, the FDR may saturate at the base error rate for larger $\alpha$. 
In such cases, the calibrated threshold can accept nearly all test samples while still satisfying the target risk constraint. 
This explains why some entries at larger risk levels have FDR values close to the base error rate and power close to one.

\paragraph{\emph{Answering Power on TriviaQA.}}
Table~\ref{tab:triviaqa_power} reports the answering power on TriviaQA. 
The results show a clear monotonic trend: as the target risk level $\alpha$ increases, CIC accepts a larger fraction of test samples. 
At strict risk levels, the method abstains more frequently to maintain reliability; at relaxed risk levels, it returns more answers and often reaches full or near-full power. 
This behavior reflects the central trade-off in selective answering: lower risk requires more abstention, while higher risk permits broader coverage.

Comparing the two UCB variants, UCB-HFD and UCB-CP exhibit different conservativeness patterns. 
UCB-CP is often more effective when the accepted subset is small, because it directly models the binomial error proportion and can produce sharper confidence bounds in low-count regimes. 
However, the relative advantage varies across models and datasets, depending on the calibration sample size, base error rate, and discriminative quality of the uncertainty estimator.

\paragraph{\emph{Answering Power on CommonsenseQA.}}
Table~\ref{tab:commonsenseqa_power} reports the answering power on CommonsenseQA. 
The same risk--coverage trade-off can be observed. 
For stronger models or settings where semantic entropy better separates correct from incorrect predictions, CIC achieves high power even under relatively strict risk levels. 
For weaker models or less discriminative uncertainty scores, the method becomes more conservative and may abstain from a larger fraction of samples. 
This is desirable in reliability-sensitive deployment, because the system should avoid answering when the calibration data cannot certify the desired error level.

Across both datasets, the empirical results support the main design principle of CIC: uncertainty scores need not be perfectly calibrated by themselves, as long as they provide a useful ranking signal. 
The confidence-interval calibration layer then converts this ranking signal into a threshold with finite-sample risk control.

\begin{figure}[!t]
    \centering
    \includegraphics[width=0.6\linewidth]{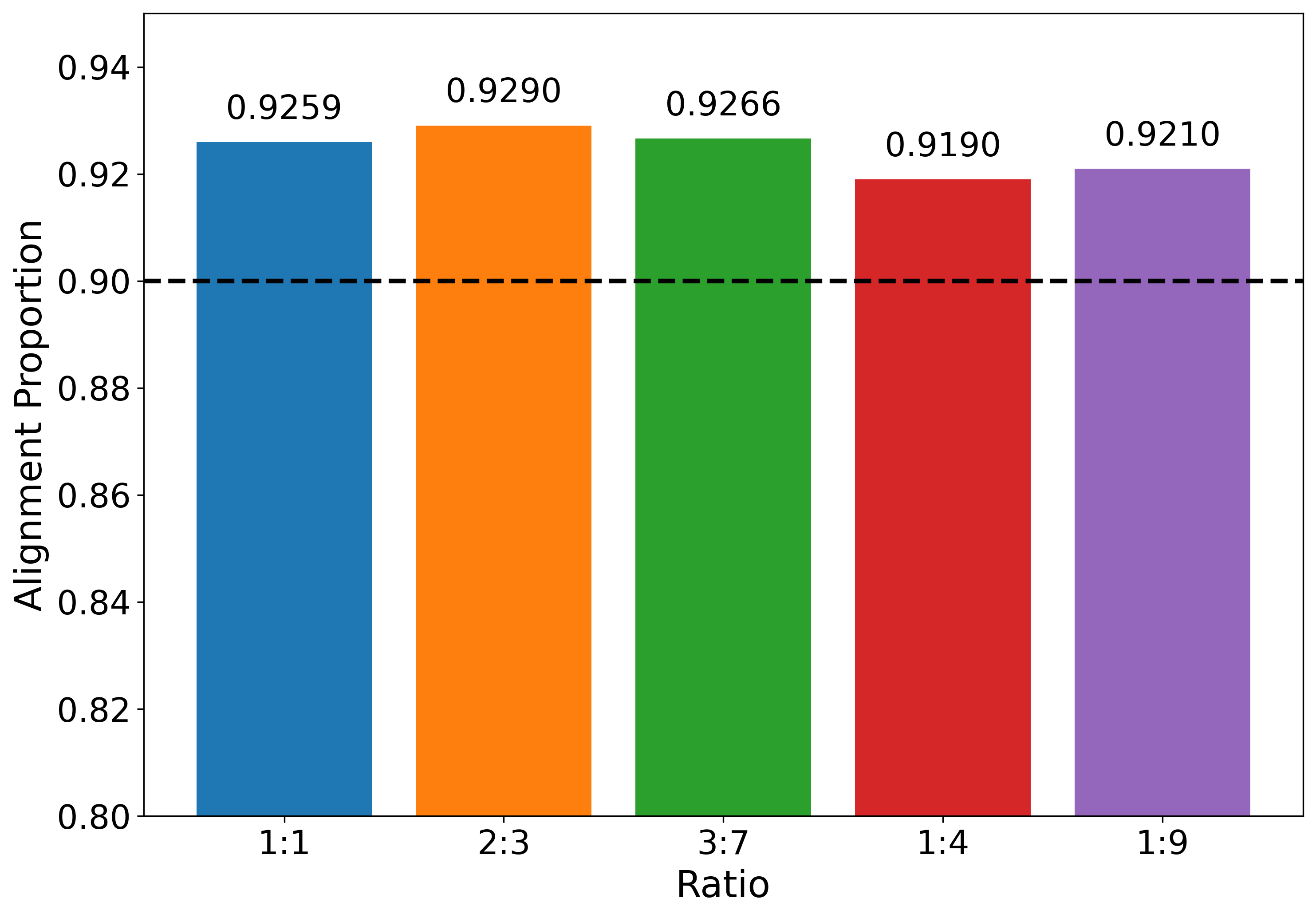}
    \caption{Sensitivity analysis under different calibration-test split ratios on CommonsenseQA using Qwen2.5-3B-Instruct. The x-axis reports the calibration:test split ratio, where 1:1, 2:3, 3:7, 1:4, and 1:9 correspond to calibration proportions of 0.5, 0.4, 0.3, 0.2, and 0.1, respectively.
The dashed line indicates the target alignment proportion of 0.90.}
    \label{fig:split_ratio}
\end{figure}

\paragraph{\emph{Sensitivity to Calibration-Test Split Ratios.}}
We further evaluate the robustness of CIC under different calibration-test split ratios.
By default, we use a 1:1 split between calibration and test data. 
To examine whether the empirical reliability is sensitive to the calibration size, we additionally vary the calibration-test ratio from 1:1 to 2:3, 3:7, 1:4, and 1:9, corresponding to calibration proportions of 0.5, 0.4, 0.3, 0.2, and 0.1, respectively.
We conduct this sensitivity study on CommonsenseQA using Qwen2.5-3B-Instruct.

As shown in Figure~\ref{fig:split_ratio}, CIC remains stable across all evaluated split ratios.
The empirical alignment proportion stays above the target level of 0.90, with values ranging from 0.9190 to 0.9290.
Although smaller calibration sets may in principle lead to wider confidence intervals and more conservative threshold selection, the observed performance does not degrade substantially in this experiment.
These results suggest that CIC is empirically robust to moderate changes in the calibration-test split ratio.

\paragraph{Discussion.}
Overall, the experiments demonstrate three key findings. 
First, CIC provides reliable control of the error rate among accepted answers across closed-ended and open-ended QA benchmarks. 
Second, the calibrated threshold naturally adapts to the user-specified risk level, yielding a smooth trade-off between reliability and answering efficiency. 
Third, infeasible cases mainly occur under very strict risk requirements, especially when the base model error rate is high or the uncertainty estimator has limited discriminative power. 
These findings are consistent with the theoretical interpretation of CIC: the method can certify reliable thresholds when the model--uncertainty pair supports the desired risk level, and otherwise correctly reports that the requested risk level is unattainable.

\section{Conclusion}

In this paper, we proposed \textbf{CIC}, a confidence-interval-based calibration framework for risk-controlled selective answering with large language models. 
Instead of assuming that heuristic uncertainty scores are intrinsically calibrated, CIC treats them as black-box ranking signals and learns a statistically certified abstention threshold from a held-out calibration set. 
By estimating the error rate among accepted responses and constructing a high-probability upper confidence bound, CIC selects the largest threshold satisfying a user-specified risk level, thereby balancing answering efficiency and deployment reliability.

Theoretically, we established finite-sample, high-probability guarantees under exchangeability, showing that the selected threshold, if feasible, controls the acceptance-conditioned error rate at the target level. 
Empirically, experiments on both closed-ended and open-ended QA benchmarks demonstrate that CIC achieves reliable risk control across multiple LLMs and uncertainty settings, while maintaining strong answering power. 
The sensitivity analysis further suggests that the framework remains stable under different calibration-test split ratios.

Overall, CIC provides a simple, model-agnostic, and practically deployable statistical layer for converting heuristic LLM uncertainty estimates into reliable selective answering policies. 
Future work may extend this framework to multi-answer generation, multi-risk constraints, adaptive threshold selection with sharper simultaneous guarantees, and more complex deployment scenarios such as retrieval-augmented generation and human-AI collaborative decision-making.

\section*{Declarations}

\subsection*{Ethics approval and consent to participate}
Not applicable.

\subsection*{Consent for publication}
Not applicable.

\subsection*{Clinical trial number}
Not applicable.

\subsection*{Funding}
The authors received no specific funding for this work.

\subsection*{Author contributions}
Sijin Dong conceived the study, developed the methodology, conducted the experiments, analyzed the results, and wrote the manuscript. Hiroyuki Shinnou supervised the study and reviewed and revised the manuscript. All authors read and approved the final manuscript.

\subsection*{Conflict of interest}
The authors declare that they have no conflict of interest or competing interests.

\subsection*{Data availability}
The datasets used in this study are publicly available benchmarks. \href{https://huggingface.co/datasets/tau/commonsense_qa/tree/main}{CommonsenseQA} and \href{https://huggingface.co/datasets/mandarjoshi/trivia_qa/tree/main}{TriviaQA} are available from their respective public repositories. Additional experimental details are provided in the manuscript.





\bibliography{sn-bibliography}

\end{document}